\documentclass[11pt]{article}

\usepackage{acl}
\usepackage{times}
\usepackage{latexsym}

\usepackage[T1]{fontenc}

\usepackage[utf8]{inputenc}

\usepackage{microtype}

\usepackage{inconsolata}

\usepackage{graphicx}
\usepackage{amssymb}
\usepackage{amsmath}
\usepackage{tabularx}
\usepackage{makecell}
\usepackage{float}
\usepackage{booktabs}
\usepackage{listings}
\usepackage{xcolor}
\usepackage{multirow}  
\usepackage{xspace}

\newcommand{\paratitle}[1]{\vspace{1.5ex}\noindent\textbf{#1}}

\newcommand{\vs}{\emph{vs.}\xspace}

\newcommand{\eg}{\emph{e.g.,}\xspace}

\newcommand{\ignore}[1]{}

\title{Squrve: A Unified and Modular Framework for Complex Real-World Text-to-SQL Tasks}

\author{ 
    Yihan Wang\textsuperscript{1} \quad
    Peiyu Liu\textsuperscript{2}\thanks{$\ $ Corresponding authors.} \quad 
    Runyu Chen\textsuperscript{2} \quad
    Jiaxing Pu\textsuperscript{3} \quad
    Wei Xu\textsuperscript{1} \\
    \textsuperscript{1}Renmin University of China \quad
    \textsuperscript{2}University of International Business and Economics \quad \\
    \textsuperscript{3}Zhejiang University of Technology \\
    \texttt{yihan3123@gmail.com} \quad
    \texttt{liupeiyustu@163.com} \quad
    \texttt{ry.chen@uibe.edu.cn} \quad \\
    \texttt{pujiaxing@zjut.edu.cn} \quad
    \texttt{weixu@ruc.edu.cn} \quad
}

\begin{document}
\maketitle
\begin{abstract}

Text-to-SQL technology has evolved rapidly, with diverse academic methods achieving impressive results. However, deploying these techniques in real-world systems remains challenging due to limited integration tools.
Despite these advances, we introduce Squrve, a unified, modular, and extensive Text-to-SQL framework designed to bring together research advances and real-world applications. Squrve first establishes a universal execution paradigm that standardizes invocation interfaces, then proposes a multi-actor collaboration mechanism based on seven abstracted effective atomic actor components. Experiments on widely adopted benchmarks demonstrate that the collaborative workflows consistently outperform the original individual methods, thereby opening up a new effective avenue for tackling complex real-world queries. The codes are available at~\href{https://github.com/Satissss/Squrve}{https://github.com/Satissss/Squrve}.

\end{abstract}

\begin{figure*}[ht]
    \centering
    \includegraphics[width=\textwidth]{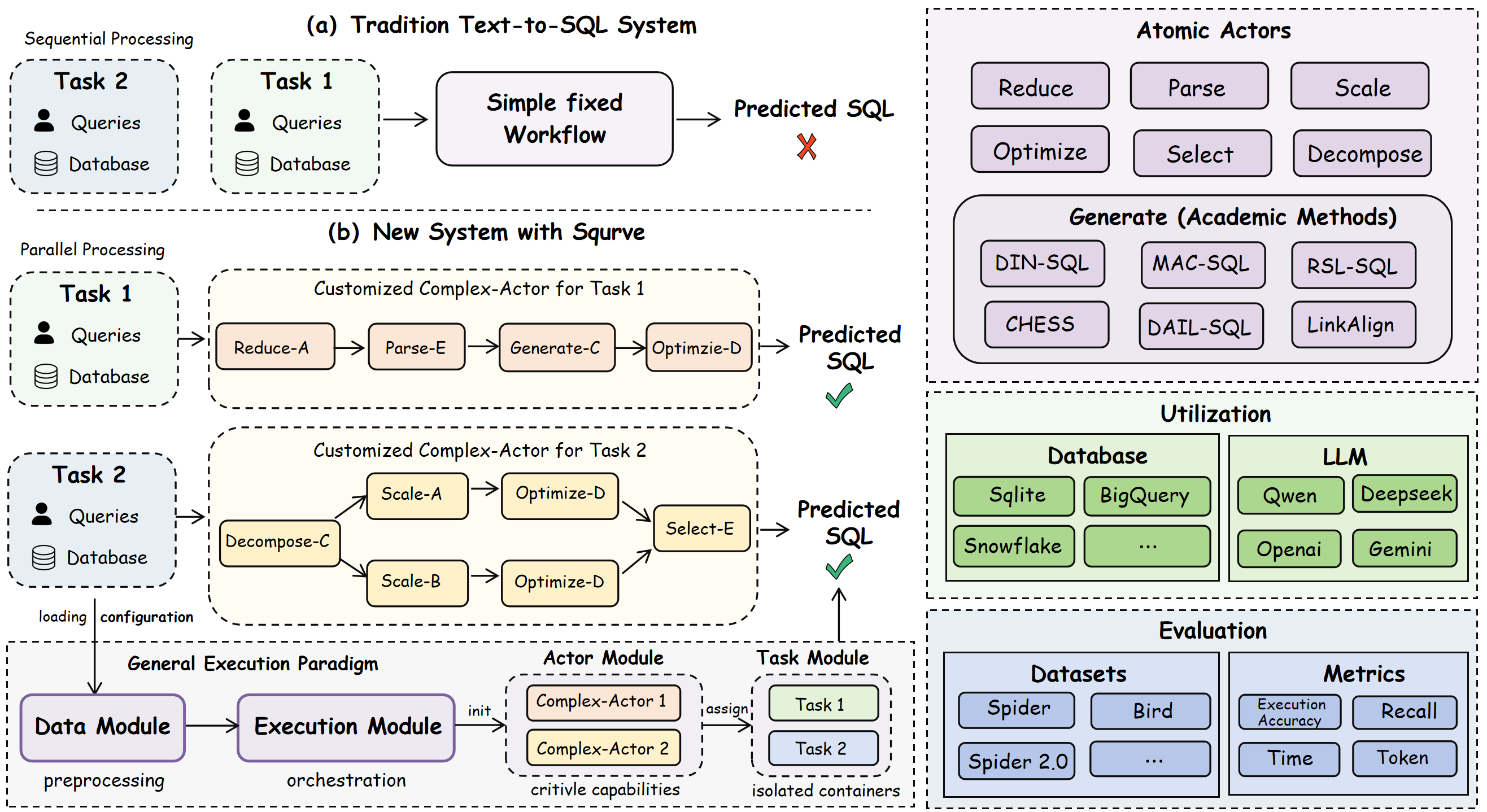}  
    \caption{The overview of the Squrve framework. Squrve unifies diverse methods under a general execution paradigm and supports task-specific multi-actor collaboration, bridging academic research and real-world systems.}
    \label{fig:overview}
\end{figure*}

\section{Introduction}\label{sec:introduction}

Text-to-SQL~\cite{liu2024survey, shi2025survey, hong2025survey} aims to translate natural language questions into accurate SQL queries, enabling non-expert users to access and manipulate data effortlessly. As an emerging querying paradigm, Text-to-SQL facilitates real-world database-driven AI agent applications, such as conversational business intelligence tools or automated decision-support assistants~\cite{wang2022proton, xu2020autoqa}. 
However, most existing studies tend to focus on specific aspects of the problem, \eg schema linking~\cite{pourreza2023dinsql}, decoding strategy~\cite{pourreza2024chase}, or data augmentation~\cite{talaei2024chess}. In contrast, real-world Text-to-SQL problems are inherently more complex, where multiple challenges often coexist. This discrepancy highlights the need for a unified framework that can systematically integrate, compare, and extend different methods, enabling complementary strengths from diverse approaches to be effectively combined and applied to practical tasks.

There are two critical challenges that hinder the development of such a unified framework. 
\emph{Challenge 1: lack of unified integration mechanism}. Existing methods are typically designed for specific settings, each with distinct data formats, input–output interfaces, or dependency requirements. This fragmentation makes it difficult to combine or switch models across diverse scenarios, resulting in limited cross-task adaptability~\cite{gan2021exploring} and poor scalability to heterogeneous real-world environments. 
\emph{Challenge 2: tackling diverse and co-occuring difficulties in real-world scenarios}. Real-world queries often involve multiple difficulties~(\eg domain adaptation and multi-table reasoning) that no single model can fully address. Integrating complementary strengths across models is therefore crucial but remains difficult without coordinated interfaces and execution pipelines. 
These challenges are often overlooked, widening the gap between academic research and real-world Text-to-SQL applications.

To address these challenges, we propose Squrve, a unified, modular, and extensive Text-to-SQL framework that systematically addresses the challenges of diverse real-world applications by fully leveraging academic advances. For Challenge 1, Squrve introduces a universal Text-to-SQL execution paradigm that standardizes invocation interfaces and parameter configurations. Through its modular architecture, Squrve reproduces and integrates existing advanced academic models, which achieves effective generalization across diverse Text-to-SQL task scenarios. For Challenge 2, Squrve abstracts and formalizes seven atomic Actor components, each representing a distinct Text-to-SQL capability validated in prior research. Building on these components, Squrve introduces a \emph{multi-actor collaboration mechanism} that effectively fuses their complementary strengths, allowing different actors to interact and cooperate when handling complex real-world queries.

To evaluate Squrve's performance, we conduct comprehensive experiments on the widely adopted Spider~\cite{yu2018spider} and Bird~
\cite{li2023bird} benchmarks, systematically assessing the execution accuracy of the generated SQL queries. 
The results indicate that Squrve faithfully reproduces baseline methods while preserving their performance, which achieves or even surpasses the originally reported results based solely on open-source models. 
Furthermore, the ensemble methods built upon multiple actors consistently outperform the original individual methods, demonstrating the effectiveness of 
compositional reasoning through the integration of complementary actors.
These findings highlight that Squrve bridges the gap between academic research and real-world applications, establishing a pathway for advancing practical development.

\section{Architecture Design}
\label{sec:workflow}

To generalize academic techniques to diverse real-world scenarios, Squrve proposes a unified, modular and extensible framework comprising four key modules.
Specifically, the \textbf{data module} standardizes different dataset formats into unified structures;  the \textbf{actor module} defines critical atomic Text-to-SQL capability and enables their collaborative composition; the \textbf{task module} serves as isolated containers that enable concurrent processing; and the \textbf{execution module} orchestrates the complete Text-to-SQL execution with global control.

\subsection{Data Module}

To standardize data operations, Squrve first defines unified structures for database schemas and query instances, following the field naming conventions of mainstream datasets such as Spider, Bird, and others. Building on this foundation, the Data module provides universal preprocessing methods that transform both user-uploaded private datasets and built-in benchmarks into standardized formats for seamless access in downstream stages. 

\paratitle{Database schema preprocessing.}  Existing datasets typically store complete large-scale database schemas in a single centralized file, which often exceeds the LLM's context window. To alleviate this limitation, this module decomposes centralized database schemas into parallel column-level files, where each file serves as a semantic unit containing complete column metadata without additional storage overhead, enabling efficient retrieval of relevant subsets via RAG methods. 

\paratitle{Query instance preprocessing.} Squrve constructs and maintains more than 200 chain-of-thought~(CoT) exemplars from existing benchmark datasets, which can be embedded and indexed for efficient vector retrieval. For new complex queries, this module retrieves semantically similar exemplars and assembles them into complete CoT prompts, effectively enhancing the quality of SQL generation. For queries that depend on external documents, this module extracts key contextual information, such as concept definitions and metric calculation formulas, using carefully designed chain-of-thought strategies.


\subsection{Actor Module}\label{sec:component}
To better handle complex real-world queries, Squrve decomposes the Text-to-SQL process into seven atomic actors, each capturing a fundamental capability that has been validated in prior research~(see Table~\ref{tab:actor_research}). Compared with complete end-to-end workflows, these actors generalize more flexibly across tasks by focusing on the core functional components. 
Building on these actors, Squrve introduces a multi-actor collaboration mechanism that integrates actors derived from different methods, enabling them to work jointly and share complementary strengths within unified workflows.
For clarity in subsequent discussion, an actor can be described as a black-box that generates the required SQL end-to-end:
\begin{equation}
    SQL = f_\text{actor}(Q, S, C|M),
\end{equation}
where $Q$ denotes user queries, $S$ denotes the database schemas, $C$ represents valuable contextual information specific to the query, and $M$ indicates the LLM backbone. Next, we will introduce the atomic actors and their collaboration mechanism in detail.

\subsubsection{Atomic Actors}
This section provides a detailed explanation of the seven atomic actors in Squrve, outlining their individual roles and functional contributions within the overall Text-to-SQL workflow.

\begin{table*}[]
    \centering
    \small
    \begin{tabular}{cccc}
    \toprule
        Actors & Methods & Key Challenges \\ \midrule
        Reduce & LinkAlign~\cite{wang2025linkalign} & large-scale and multi-database \\
        Parse  & LinkAlign; RSL-SQL~\cite{cao2024rsl} & ambiguous queries and redundant schemas \\
        Generate & DIN-SQL; CHESS~\cite{ talaei2024chess} & efficient and high-quality SQL generation   \\
        Decompose & DIN-SQL; MAC-SQL~\cite{wang2024macsql} & Chain-of-Thought for complex queries \\
        Scale  & CHESS; CHASE-SQL~\cite{pourreza2024chase} & diverse and high-quality decoding strategies \\
        Optimize & CHASE-SQL; OpenSearch~\cite{xie2025opensearch};   & effective and broader database feedback  \\
        Select & OpenSearch; MCS-SQL~\cite{lee2024mcs};  & accurate gold SQL identification \\
    \bottomrule
    \end{tabular}
    \caption{Summary of the prior methods that each Actor in Squrve builds upon, and the corresponding challenges highlighted in earlier research.}
    \label{tab:actor_research}
\end{table*}

\paratitle{Reduce actor.} This component eliminates redundant schemas from large-scale databases that may exceed LLM context windows. 
By retaining only the relevant subsets, the reduce actor allows subsequent stages to operate on compact and focused schema representations. 

\paratitle{Parse actor.} This component performs schema linking by extracting tables and columns from candidate schemas that are potentially required for SQL generation. By providing attention signals, the parse actor enables easier focus on critical elements during subsequent SQL generation. 

\paratitle{Generate actor.} This component generates complete SQL statements, encapsulating existing end-to-end academic methods. The generator actor can be applied directly to Text-to-SQL tasks in straightforward scenarios. 

\paratitle{Decompose actor.} This component decomposes complex queries into multiple logically progressive sub-questions and generates SQL statements for each. By simulating chain-of-thought~(CoT) reasoning, the decompose actor provides interpretable reasoning pathways for complex queries. 

\paratitle{Scale actor.} This component generates diverse high-quality SQL candidates to increase the probability of covering the gold SQL. Through leveraging abundant inference-time computation, the scale actor improves generation quality and overcomes the limitations of the single-decoding strategy. 


\paratitle{Optimize actor.} This component leverages environmental feedback (\eg database errors or  results) to refine the quality of the generated SQL queries. Through feedback signals, the optimize actor can rapidly locate syntax errors or identify discrepancies between retrieval results and user intent. 

\paratitle{Select actor.} This component selects the optimal SQL statement from multiple candidates, typically in collaboration with the scale actor. By filtering out suboptimal candidates, the select actor enables broad explorations of generated SQL to converge toward the final output. 

\subsubsection{Complex actors}
\label{sec-complexactors}
To support scalable integration of atomic actors, Squrve introduces two composition strategies, \textbf{Pipeline} and \textbf{Tree}, which serve as structural templates with plug-and-socket connection interfaces. {These strategies wrap multiple actors into a single composite actor, which can be recursively applied to construct complex workflows.}


\paratitle{Pipeline.} 
The Pipeline strategy connects multiple actors in a sequential manner, where the output of each stage serves as the input to the next. 
This design enables multi-step transformations through staged processing, allowing intermediate results to be refined progressively along the workflow.

\paratitle{Tree.} 
The Tree strategy dispatches a shared root input to multiple child actors for parallel processing and then merges their outputs into a unified result. 
This structure supports ensemble-style execution, where different processing pathways contribute complementary interpretations or candidate SQLs that collectively improve overall performance.

\paratitle{Identified combinations.}
To explore effective collaboration patterns, we evaluate multiple actor combinations and identify two favorable ensembles~(see Section~\ref{sec-setup}). Specifically, we randomly sample 50 instances from the Spider and BIRD datasets, and use Qwen3-turbo as the backbone model to generate predictions. All results are summarized in Table~\ref{tab:ex_results}, based on which we construct two high-performing multi-actor workflows that serve as the foundation for subsequent formal experiments.

\subsection{Task Module}

The task module coordinates the execution of Text-to-SQL tasks by linking actors, data, and runtime environments under a unified interface.  
Specifically, it includes an \textbf{actor interface}, used to perform the core Text-to-SQL conversion process; {a \textbf{dataset interface}, used to load preprocessed task-specific inputs such as user queries and database schemas through data module's interface}; and a \textbf{runtime controller}, used to manage execution flow {(\eg concurrency control)} and collect generated SQL outputs.  
Together, these components provide an isolated and concurrent runtime environment, ensuring efficient task scheduling and preventing interference across different executions.

\subsection{Execution Module}

The execution module serves as the central controller of Squrve, responsible for coordinating the entire Text-to-SQL workflow—from system initialization to final evaluation.  
Specifically, it includes a \textbf{configuration manager}, used to load system settings and initialize all task containers; an \textbf{execution controller}, used to launch, monitor, and manage concurrent task executions while handling logging and exception recovery; and an \textbf{evaluation unit}, used to aggregate results, compute metrics such as execution accuracy, and generate evaluation reports.  
Together, these components ensure stable large-scale execution and reproducible performance assessment across diverse Text-to-SQL scenarios.

\section{Usage Examples}\label{sec:usage}
This section illustrates the usage of Squrve, covering general startup and invocation, running baseline methods on benchmark datasets, and composing customized workflows from atomic components.

\subsection{General Startup Demonstration}\label{sec:startup}
Overall, Squrve offers a unified interface that executes diverse Text-to-SQL workflows through minimal user configuration. 
As illustrated in Appendix~\ref{app:example}, users can run different methods with only a few lines of code. 
The framework automatically handles data loading, execution, and evaluation, producing complete results including predicted SQLs, query outputs, and evaluation metrics. 
This unified interface simplifies the use of advanced Text-to-SQL methods, then users can execute and compare different models with a single line of code, without the need to modify separate scripts or pipelines. 
In future releases, we plan to further enrich the configuration system, enabling users to flexibly test various benchmarks and baseline methods in Section~\ref{usage:baselines}. 
Moreover, the configuration will support fine-grained control over Actor settings, allowing users to define customized composition strategies as detailed in Section~\ref{sec-customize}.

\subsection{Running Existing Baselines}\label{usage:baselines}


{Unlike existing systems that rely on fixed, task-specific algorithms, Squrve integrates a variety of advanced open-source methods that can be flexibly applied to diverse Text-to-SQL scenarios.
For a given dataset, this can be easily achieved through the following steps:}

\begin{itemize}
    \item[$\bullet$] \textbf{Formatting the dataset.} For existing widely used benchmarks, Squrve has pre-unified their formats and maintains them in the 'benchmark' directory, requiring no additional operations. For proprietary datasets, users need to preprocess them into valid formats following Squrve's documentation. Additionally, databases with access control (\eg BigQuery) require users to prepare authentication credentials in advance.

    \item[$\bullet$] \textbf{Generating the configuration.} Squrve provides flexible approaches to generate configurations, including configuration files, command-line arguments, or direct code specification. Users typically need to specify the dataset path, method name, task definition, and other configuration settings.

    \item[$\bullet$] \textbf{Starting the execution process.} Once the configuration is prepared, users can directly start the execution process. Squrve automatically handles dataset initialization, task execution, and output saving.

    \item[$\bullet$] \textbf{Evaluating and organizing the results.} After task completion, users can evaluate the performance using multiple evaluation metrics, such as execution accuracy, and further analyze results through visualization.
\end{itemize}

\subsection{Customize Implementations}
\label{sec-customize}
{For challenging tasks, users may need to integrate advantages from different methods and design customized workflows. This can be achieved by replacing method selection with workflow design in the generating the configuration} step:

\begin{itemize}
    \item[$\bullet$] \textbf{Selecting the atomic actors.} Atomic actors are the fundamental components, like the bricks that form a building. Therefore, users first select the necessary atomic actors for their workflow, such as DINSQLParser for the schema linking step.

    \item[$\bullet$] \textbf{Designing the customized workflow.} Users can scale up the final performance through two composition strategies: serially chaining different actors or composing complementary actors in parallel. Squrve represents such topological structures using nested lists. For more advanced usage, users can construct more sophisticated models based on previously defined simple workflows.
\end{itemize}

\begin{table*}[h]
\small
\centering
\resizebox{0.55\textwidth}{!}{
\begin{tabular}{lccccc}
\toprule
\multirow{2}{*}{\textbf{Methods}} & \multirow{2}{*}{\textbf{Individual}} & \multicolumn{2}{c}{\textbf{EX(Spider-dev)}} & \multicolumn{2}{c}{\textbf{EX(Bird-dev)}} \\
\cmidrule(lr){3-4} \cmidrule(lr){5-6}
 & & Origin & Ours & Origin & Ours \\
\midrule
DIN-SQL      & $\checkmark$ & 81.7 & 82.5 & 59.4 & 60.1 \\
CHESS        & $\checkmark$ & 85.6 & 84.8 & 64.6 & 63.1 \\
MAC-SQL      & $\checkmark$ & 82.5 & 86.0 & \textbf{67.7} & 67.0 \\
RSL-SQL      & $\checkmark$ & 86.5 & 87.3 & 67.0 & 68.0 \\
LinkAlign    & $\checkmark$ & \textbf{87.2} & 86.5 & 64.0 & 65.5 \\\midrule
Ensemble-1   & × & N/A  & 88.7 & N/A  & 67.3 \\
Ensemble-2   & × & N/A  & \textbf{90.8} & N/A  & \textbf{70.0} \\
\bottomrule
\end{tabular}
}
\caption{Comparison between individual baseline methods and our variants on Qwen-3 backbone. The \emph{Origin} reports results obtained by executing the original repository code; \emph{Ours} presents the reproduced performance under Squrve framework. N/A indicates our proposed variants have no results from original code execution.}
\label{tab:main_result}
\end{table*}

\section{Experiment}\label{sec:experiment}

\subsection{Experiment Setup}
\label{sec-setup}

\paratitle{Datasets.} We conduct performance evaluations on three widely adopted Text-to-SQL benchmarks, Spider~\cite{yu2018spider}, BIRD~\cite{li2023bird}, and AmbiDB~\cite{wang2025linkalign}. Spider is a large-scale, cross-domain dataset for assessing model generalization across diverse database schemas. BIRD features larger databases, noisier data, and more complex queries requiring external knowledge. AmbiDB introduces enhanced ambiguity through multi-database settings, requiring to identify target database without prior specification.

\paratitle{Baselines.} Squrve integrates multiple state-of-the-art baselines: DIN-SQL employs decomposed in-context learning with self-correction; CHESS-SQL leverages context mining for effective SQL synthesis; MAC-SQL~\cite{wang2024macsql} utilizes multi-agent collaboration to identify relevant schema subsets; RSL-SQL~\cite{cao2024rsl} adopts diverse strategies to enhance schema linking and SQL generation robustness; and LinkAlign~\cite{wang2025linkalign} addresses schema linking challenges in large-scale, multi-database environments.

\paratitle{Variants.} We identify two promising and competitive multi-actor ensemble workflows through exploring different combination strategies in a small-scale subset, as detailed in Section~\ref{sec-complexactors}. 
We further compare the performance of these variants with individual baselines through experiments.

\begin{enumerate}
  \item \textbf{Ensemble-1}: 
  \texttt{LinkAlignParser + RSLSQLScaler + CHESSSelector + MACSQLOptimizer}
  \item \textbf{Ensemble-2}: 
  \texttt{RSLSQLBiDirParser + MACSQLDecompose + ChessScaler + CHESSSelector + LinkAlignOptimizer}
\end{enumerate}

\paratitle{Evaluation Metrics.} 
We evaluate end-to-end Text-to-SQL performance using Execution Accuracy (EX), which compares the execution results of predicted and gold SQL queries. For schema linking, Recall and Precision are used to assess the accuracy of relevant schema element retrieval.

\paratitle{Implementations.} We use Qwen-3 as the LLM backbone. Building upon the paradigm introduced in Section~\ref{sec:workflow}, Squrve enables seamless extensibility of benchmark datasets and baseline methods along two orthogonal dimensions. We can freely combine method-dataset pairs and execute task batches concurrently for improved efficiency.

\vspace{-0.05cm}
\subsection{Main Results}
\paragraph{Results of existing methods}
As shown in Table~\ref{tab:main_result}, Squrve successfully reproduces existing Text-to-SQL baselines under the same LLM backbone with performance closely aligned to their originally executed results. Specifically, our reproduced results achieve comparable or even superior execution accuracy across both benchmarks. For instance, DIN-SQL achieves 82.5\% on Spider-dev (82.5\% \vs original 81.7\%) and improves to 60.1\% on Bird-dev (60.1\% \vs original 59.4\%). Overall, this faithful reproduction establishes a reliable foundation for fair comparison.

\vspace{-0.2cm}
\paragraph{{Results of our variants.}}
Building upon the reproduced components, we further demonstrate the extensibility and effectiveness of the Multi-Actor collaboration mechanism through evaluation on two variants. As shown in Table~\ref{tab:main_result}, both ensemble methods substantially outperform all individual baselines in both benchmarks. In particular, Ensemble-2 achieved the best performance on both benchmarks with execution accuracies of 90.8\% and 70.0\%, improving 4.0\% and 2.9\% over the strongest individual method, respectively. These results demonstrate the potential of Multi-Actor collaboration to address complex queries by integrating complementary method strengths.





\subsection{Parallel Component Analysis}
While serially chaining multiple atomic actors scales up the final performance, parallel composition of complementary homogeneous actors orthogonal to the workflow can effectively enhance intermediate stages, such as schema linking, without efficiency loss. As shown in Table~\ref{tab:parallel_actor}, all the parallel ensemble methods outperform individual methods in recall, which is typically prioritized in schema linking evaluation.  Notably, two-actor parallelization strikes the optimal balance, as additional actors provide diminishing marginal gains while introducing computational overhead. This finding offers important insights for effectively harnessing multi-actor collaboration.

\begin{table}[ht]
\centering
\small
\begin{tabularx}{\columnwidth}{@{}X*{4}{c}@{}}
\toprule
\multirow{2}{*}{\textbf{Methods}} & \multicolumn{2}{c}{\textbf{AmbiDB}} & \multicolumn{2}{c}{\textbf{Bird-dev}} \\
\cmidrule(lr){2-3} \cmidrule(lr){4-5}
 & Recall & Precision & Recall & Precision \\
\midrule
LinkAlign & 49.0 & 51.5 & 76.9 & 81.5 \\
RSL-SQL & 54.5 & 8.2 & 82.4 & 35.3 \\
DIN-SQL & 21.6 & 23.0 & 59.8 & 34.2 \\
Link.+RSL. & 61.8 & 13.9 & 92.4 & 40.0 \\
Link.+DIN. & 55.3 & 42.4 & 81.2 & 58.1 \\
RSL.+DIN. & 64.5 & 8.1 & 87.1 & 29.0 \\
Link.+RSL.+DIN. & 62.3 & 7.3 & 93.1 & 28.7 \\
\bottomrule
\end{tabularx}
\caption{Parallel actors performance comparison.}
\label{tab:parallel_actor}
\end{table}

\vspace{-0.5cm}
\section{Conclusion}\label{sec:conclusion}
This paper introduces Squrve, a unified, modular, and extensive Text-to-SQL framework that bridges the gap between academic researches and real-world applications. Through a universal workflow paradigm and modular architecture, Squrve decouples capabilities from usages, enabling researchers to seamlessly integrate, replace, and reuse technical components. Experiments validate the effectiveness, demonstrating that ensemble workflows outperform single end-to-end models. We believe Squrve will serve as a foundational platform, accelerating real-world Text-to-SQL applications.


\bibliography{custom}

\begin{thebibliography}{40}
\providecommand{\natexlab}[1]{#1}

\bibitem[{Cao et~al.(2024)Cao, Zheng, Fan, Zhang, Chen, and Bai}]{cao2024rsl}
Zhenbiao Cao, Yuanlei Zheng, Zhihao Fan, Xiaojin Zhang, Wei Chen, and Xiang Bai. 2024.
\newblock Rsl-sql: Robust schema linking in text-to-sql generation.
\newblock \emph{arXiv preprint arXiv:2411.00073}.

\bibitem[{Choi et~al.(2021)Choi, Shin, Kim, and Shin}]{choi2021ryansql}
DongHyun Choi, Min~C Shin, EungGyun Kim, and Dong~Ryeol Shin. 2021.
\newblock Ryansql: Recursively applying sketch-based slot fillings for complex text-to-sql in cross-domain databases.

\bibitem[{Deng et~al.(2025)Deng, Ramachandran, Xu, Hu, Yao, Datta, and Zhang}]{deng2025reforce}
Michael Deng, Akshara Ramachandran, Cheng Xu, Lyuzhou Hu, Zhoujun Yao, Anindya Datta, and Heng Zhang. 2025.
\newblock Reforce: A text-to-sql agent with self-refinement, format restriction, and column exploration.
\newblock In \emph{ICLR 2025 The VerifAI Workshop}.

\bibitem[{Deng et~al.(2021)Deng, Awadallah, Meek, Polozov, Sun, and Richardson}]{deng2021structure}
Xiang Deng, Ahmed~Hassan Awadallah, Christopher Meek, Oleksandr Polozov, Huan Sun, and Matthew Richardson. 2021.
\newblock Structure-grounded pretraining for text-to-sql.
\newblock In \emph{North American Chapter of the Association for Computational Linguistics: Human Language Technologies}.

\bibitem[{Devlin et~al.(2019)Devlin, Chang, Lee, and Toutanova}]{devlin2019bert}
Jacob Devlin, Ming-Wei Chang, Kenton Lee, and Kristina Toutanova. 2019.
\newblock Bert: Pre-training of deep bidirectional transformers for language understanding.
\newblock In \emph{North American Chapter of the Association for Computational Linguistics: Human Language Technologies}.

\bibitem[{Dong et~al.(2023)Dong, Zhang, Ge, Mao, Gao, Lin, Lou et~al.}]{dong2023c3}
Xuemei Dong, Chao Zhang, Yuhang Ge, Yuren Mao, Yunjun Gao, Jinshu Lin, Dongfang Lou, and 1 others. 2023.
\newblock C3: Zero-shot text-to-sql with chatgpt.

\bibitem[{Gan et~al.(2021{\natexlab{a}})Gan, Chen, Huang, Purver, Woodward, Xie, and Huang}]{gan2021synonym}
Yujian Gan, Xinyun Chen, Qiuping Huang, Matthew Purver, John~R Woodward, Jinxia Xie, and Pengsheng Huang. 2021{\natexlab{a}}.
\newblock Towards robustness of text-to-sql models against synonym substitution.
\newblock In \emph{Association for Computational Linguistics and International Joint Conference on Natural Language Processing}.

\bibitem[{Gan et~al.(2021{\natexlab{b}})Gan, Chen, and Purver}]{gan2021exploring}
Yujian Gan, Xinyun Chen, and Matthew Purver. 2021{\natexlab{b}}.
\newblock Exploring underexplored limitations of cross-domain text-to-sql generalization.
\newblock In \emph{Empirical Methods in Natural Language Processing (EMNLP)}.

\bibitem[{Gao et~al.(2024)Gao, Wang, Li, Sun, Qian, Ding, and Zhou}]{gao2024dail}
Dawei Gao, Haibin Wang, Yaliang Li, Xiuyu Sun, Yue Qian, Bolin Ding, and Jingren Zhou. 2024.
\newblock Text-to-sql empowered by large language models: A benchmark evaluation.
\newblock In \emph{International Conference on Very Large Data Bases}.

\bibitem[{Hong et~al.(2025)Hong, Yuan, Zhang, Chen, Dong, Huang, and Huang}]{hong2025survey}
Zijin Hong, Zheng Yuan, Qinggang Zhang, Hao Chen, Junnan Dong, Feiran Huang, and Xiao Huang. 2025.
\newblock Next-generation database interfaces: A survey of llm-based text-to-sql.
\newblock \emph{arXiv preprint arXiv:2406.08426}.

\bibitem[{Hui et~al.(2021)Hui, Shi, Geng, Li, Li, Sun, and Zhu}]{hui2021schema}
Binyuan Hui, Xiang Shi, Ruiying Geng, Binhua Li, Yongbin Li, Jian Sun, and Xiaodan Zhu. 2021.
\newblock Improving text-to-sql with schema dependency learning.

\bibitem[{Kou et~al.(2024)Kou, Hu, He, Deng, and Zhang}]{kou2024cllms}
Siqi Kou, Lanxiang Hu, Zhezhi He, Zhijie Deng, and Hao Zhang. 2024.
\newblock Cllms: Consistency large language models.
\newblock In \emph{International Conference on Machine Learning}.

\bibitem[{Lee et~al.(2024)Lee, Park, Kim, and Park}]{lee2024mcs}
Dongjun Lee, Choongwon Park, Jaehyuk Kim, and Heesoo Park. 2024.
\newblock Mcs-sql: Leveraging multiple prompts and multiple-choice selection for text-to-sql generation.
\newblock \emph{arXiv preprint arXiv:2405.07467}.

\bibitem[{Lei et~al.(2025)Lei, Chen, Ye, Cao, Shin, SU, SUO, Gao, Hu, Yin et~al.}]{lei2024spider2}
Fangyu Lei, Jixuan Chen, Yuxiao Ye, Ruisheng Cao, Dongchan Shin, Hongjin SU, Zhaoqing SUO, Hongcheng Gao, Wenjing Hu, Pengcheng Yin, and 1 others. 2025.
\newblock Spider 2.0: Evaluating language models on real-world enterprise text-to-sql workflows.
\newblock In \emph{International Conference on Learning Representations}.

\bibitem[{Li and Jagadish(2014)}]{li2009rule}
Fei Li and HV~Jagadish. 2014.
\newblock Constructing an interactive natural language interface for relational databases.
\newblock In \emph{International Conference on Very Large Data Bases}.

\bibitem[{Li et~al.(2023{\natexlab{a}})Li, Zhang, Li, and Chen}]{li2023resdsql}
Haoyang Li, Jing Zhang, Cuiping Li, and Hong Chen. 2023{\natexlab{a}}.
\newblock Resdsql: Decoupling schema linking and skeleton parsing for text-to-sql.
\newblock In \emph{Conference on Artificial Intelligence}.

\bibitem[{Li et~al.(2023{\natexlab{b}})Li, Hui, Cheng, Qin, Ma, Huo, Huang, Du, Si, and Li}]{li2023graphix}
Jinyang Li, Binyuan Hui, Reynold Cheng, Bowen Qin, Chenhao Ma, Nan Huo, Fei Huang, Wenyu Du, Luo Si, and Yongbin Li. 2023{\natexlab{b}}.
\newblock Graphix-t5: Mixing pre-trained transformers with graph-aware layers for text-to-sql parsing.
\newblock In \emph{Conference on Artificial Intelligence}.

\bibitem[{Li et~al.(2023{\natexlab{c}})Li, Hui, Qu, Yang, Li, Li, Wang, Qin, Geng, Huo et~al.}]{li2023bird}
Jinyang Li, Binyuan Hui, Ge~Qu, Jiaxi Yang, Binhua Li, Bowen Li, Bailin Wang, Bowen Qin, Ruiying Geng, Nan Huo, and 1 others. 2023{\natexlab{c}}.
\newblock Can llm already serve as a database interface? a big bench for large-scale database grounded text-to-sqls.
\newblock In \emph{Advances in Neural Information Processing Systems}.

\bibitem[{Liu et~al.(2024)Liu, Shen, Li, Ma, Jiang, Zhang, Fan, Li, Tang, and Luo}]{liu2024survey}
Xinyu Liu, Shuyu Shen, Boyan Li, Peixian Ma, Runzhi Jiang, Yuxin Zhang, Ju~Fan, Guoliang Li, Nan Tang, and Yuyu Luo. 2024.
\newblock A survey of nl2sql with large language models: Where are we, and where are we going?
\newblock \emph{arXiv preprint arXiv:2408.05109}.

\bibitem[{Liu et~al.(2019)Liu, Ott, Goyal, Du, Joshi, Chen, Levy, Lewis, Zettlemoyer, and Stoyanov}]{liu2019roberta}
Yinhan Liu, Myle Ott, Naman Goyal, Jingfei Du, Mandar Joshi, Danqi Chen, Omer Levy, Mike Lewis, Luke Zettlemoyer, and Veselin Stoyanov. 2019.
\newblock Roberta: A robustly optimized bert pretraining approach.
\newblock \emph{arXiv preprint arXiv:1907.11692}.

\bibitem[{Pi et~al.(2022)Pi, Wang, Gao, Guo, Li, and Lou}]{pi2022adversarial}
Xinyu Pi, Bing Wang, Yan Gao, Jiawei Guo, Zhoujun Li, and Jian-Guang Lou. 2022.
\newblock Towards robustness of text-to-sql models against natural and realistic adversarial table perturbation.
\newblock In \emph{Association for Computational Linguistics}.

\bibitem[{Pourreza et~al.(2025)Pourreza, Li, Sun, Chung, Talaei, Kakkar, Gan, Saberi, Ozcan, and Arik}]{pourreza2024chase}
Mohammadreza Pourreza, Haiyang Li, Ruoxi Sun, Yeounoh Chung, Shayan Talaei, Gaurav~T Kakkar, Yu~Gan, Amin Saberi, Fatma Ozcan, and Sercan~O Arik. 2025.
\newblock Chase-sql: Multi-path reasoning and preference optimized candidate selection in text-to-sql.
\newblock In \emph{International Conference on Learning Representations}.

\bibitem[{Pourreza and Rafiei(2023)}]{pourreza2023dinsql}
Mohammadreza Pourreza and Davood Rafiei. 2023.
\newblock Din-sql: Decomposed in-context learning of text-to-sql with self-correction.
\newblock In \emph{Advances in Neural Information Processing Systems}.

\bibitem[{Shi et~al.(2025)Shi, Tang, Zhang, Zhang, and Yang}]{shi2025survey}
Liang Shi, Zhengju Tang, Nan Zhang, Xiaotong Zhang, and Zhi Yang. 2025.
\newblock A survey on employing large language models for text-to-sql tasks.
\newblock \emph{ACM Computing Surveys}, 58(2):1--37.

\bibitem[{Talaei et~al.(2024)Talaei, Pourreza, Chang, Mirhoseini, and Saberi}]{talaei2024chess}
Shayan Talaei, Mohammadreza Pourreza, Yu-Chen Chang, Azalia Mirhoseini, and Amin Saberi. 2024.
\newblock Chess: Contextual harnessing for efficient sql synthesis.
\newblock \emph{arXiv preprint arXiv:2405.16755}.

\bibitem[{Wang et~al.(2024)Wang, Ren, Yang, Liang, Bai, Chai, Yan, Zhang, Yin, Sun et~al.}]{wang2024macsql}
Bing Wang, Changyu Ren, Jian Yang, Xinnian Liang, Jiaqi Bai, Linzheng Chai, Zhao Yan, Qian-Wen Zhang, Di~Yin, Xipeng Sun, and 1 others. 2024.
\newblock Mac-sql: A multi-agent collaborative framework for text-to-sql.
\newblock In \emph{International Conference on Computational Linguistics}.

\bibitem[{Wang et~al.(2022{\natexlab{a}})Wang, Qin, Hui, Li, Yang, Wang, Li, Sun, Huang, Si, and Li}]{wang2022proton}
Lihan Wang, Bowen Qin, Binyuan Hui, Bowen Li, Min Yang, Bailin Wang, Binhua Li, Jian Sun, Fei Huang, Luo Si, and Yongbin Li. 2022{\natexlab{a}}.
\newblock Proton: Probing schema linking information from pre-trained language models for text-to-sql parsing.
\newblock In \emph{Conference on Knowledge Discovery and Data Mining (KDD)}.

\bibitem[{Wang et~al.(2023)Wang, Lin, Han, Sun, Chen, Wang, and Zeng}]{wang2023dbcopilot}
Tianshu Wang, Hongyu Lin, Xianpei Han, Le~Sun, Xiaoyang Chen, Hao Wang, and Zhenyu Zeng. 2023.
\newblock Dbcopilot: Scaling natural language querying to massive databases.
\newblock \emph{arXiv preprint arXiv:2312.03463}.

\bibitem[{Wang et~al.(2022{\natexlab{b}})Wang, Wei, Schuurmans, Le, Chi, Narang, Chowdhery, and Zhou}]{wang2022self}
Xuezhi Wang, Jason Wei, Dale Schuurmans, Quoc Le, Ed~Chi, Sharan Narang, Aakanksha Chowdhery, and Denny Zhou. 2022{\natexlab{b}}.
\newblock Self-consistency improves chain of thought reasoning in language models.
\newblock In \emph{ICLR}.

\bibitem[{Wang et~al.(2025)Wang, Liu, and Yang}]{wang2025linkalign}
Yihan Wang, Peiyu Liu, and Xin Yang. 2025.
\newblock Linkalign: Scalable schema linking for real-world large-scale multi-database text-to-sql.
\newblock \emph{arXiv preprint arXiv:2503.18596}.

\bibitem[{Wei et~al.(2022)Wei, Wang, Schuurmans, Bosma, Ichter, Xia, Chi, Le, and Zhou}]{wei2022chain}
Jason Wei, Xuezhi Wang, Dale Schuurmans, Maarten Bosma, Brian Ichter, Fei Xia, Ed~Chi, Quoc Le, and Denny Zhou. 2022.
\newblock Chain-of-thought prompting elicits reasoning in large language models.
\newblock In \emph{NeurIPS}.

\bibitem[{Xie et~al.(2025)Xie, Xu, Zhao, and Guo}]{xie2025opensearch}
Xiangjin Xie, Guangwei Xu, Lingyan Zhao, and Ruijie Guo. 2025.
\newblock Opensearch-sql: Enhancing text-to-sql with dynamic few-shot and consistency alignment.
\newblock \emph{Proceedings of the ACM on Management of Data}, 3(3):1--24.

\bibitem[{Xu et~al.(2020)Xu, Semnani, Campagna, and Lam}]{xu2020autoqa}
Silei Xu, Sina Semnani, Giovanni Campagna, and Monica Lam. 2020.
\newblock Autoqa: From databases to qa semantic parsers with only synthetic training data.
\newblock In \emph{Empirical Methods in Natural Language Processing (EMNLP)}.

\bibitem[{Xue et~al.(2024)Xue, Jiang, Shi, Cheng, Chen, Yang, Zhang, He, Zhang, Wei et~al.}]{xue2024dbgpt}
Siqiao Xue, Caigao Jiang, Wenhui Shi, Fangyin Cheng, Keting Chen, Hongjun Yang, Zhiping Zhang, Jianshan He, Hongyang Zhang, Ganglin Wei, and 1 others. 2024.
\newblock Db-gpt: Empowering database interactions with private large language models.

\bibitem[{Yin et~al.(2020)Yin, Neubig, Yih, and Riedel}]{yin2020tabert}
Pengcheng Yin, Graham Neubig, Wen-tau Yih, and Sebastian Riedel. 2020.
\newblock Tabert: Pretraining for joint understanding of textual and tabular data.

\bibitem[{Yu et~al.(2023)Yu, Chang, Wang, Dong, Pan, Zhu, Li, Lan, Zhang, Jiang et~al.}]{yu2023drspider}
Tao Yu, Shuaichen Chang, Jiaqi Wang, Ming Dong, Linyong Pan, Henghui Zhu, Abishek Li, Wenting Lan, Shuaigi Zhang, Jiarong Jiang, and 1 others. 2023.
\newblock Dr. spider: A diagnostic evaluation benchmark towards text-to-sql robustness.
\newblock In \emph{International Conference on Learning Representations (ICLR)}.

\bibitem[{Yu et~al.(2018)Yu, Zhang, Yang, Yasunaga, Wang, Li, Ma, Li, Yao, Roman et~al.}]{yu2018spider}
Tao Yu, Rui Zhang, Kai Yang, Michihiro Yasunaga, Dongxu Wang, Zifan Li, James Ma, Irene Li, Qingning Yao, Shanelle Roman, and 1 others. 2018.
\newblock Spider: A large-scale human-labeled dataset for complex and cross-domain semantic parsing and text-to-sql task.
\newblock In \emph{Empirical Methods in Natural Language Processing}.

\bibitem[{Zhang et~al.(2024)Zhang, Ye, Du, Hu, Li, Yang, Chi, Liu, Zhao, Li et~al.}]{zhang2024benchmarking}
Bin Zhang, Yuxiao Ye, Guosheng Du, Xiaoxu Hu, Zhishuai Li, Sun Yang, Haoxuan Chi, Rui Liu, Zichen Zhao, Zhi Li, and 1 others. 2024.
\newblock Benchmarking the text-to-sql capability of large language models: A comprehensive evaluation.
\newblock \emph{arXiv preprint arXiv:2403.02951}.

\bibitem[{Zhang et~al.(2023)Zhang, Yu, Hashimoto, Lewis, Yih, Fried, and Wang}]{zhang2023coder}
Tianyi Zhang, Tao Yu, Tatsunori~B Hashimoto, Mike Lewis, Wen-tau Yih, Daniel Fried, and Sida~I Wang. 2023.
\newblock Coder reviewer reranking for code generation.
\newblock In \emph{International Conference on Machine Learning}.

\bibitem[{Zhong et~al.(2022)Zhong, Wang, Gao, Guo, Li, and Lou}]{zhong2022adveta}
Ruiqi Zhong, Bing Wang, Yan Gao, Jiaqi Guo, Zhan Li, and Jian-Guang Lou. 2022.
\newblock Towards robustness of text-to-sql models against natural and realistic adversarial table perturbation.
\newblock In \emph{Proceedings of the 60th Annual Meeting of the Association for Computational Linguistics (Volume 1: Long Papers)}, pages~{}--{}. ACL.

\end{thebibliography}

\appendix
\begin{table*}[htbp]
\small
\centering
\begin{tabular}{ccc}
\hline
\textbf{Methods} & \textbf{EX (Spider)} & \textbf{EX (Bird)} \\
\hline
DIN-SQL & 65.9 & 47.5 \\
CHESS-SQL & 65.2 & 38.6 \\
MAC-SQL & 73.5 & 50.0 \\
DAIL-SQL & 65.9 & 53.1 \\
RSL-SQL & 70.8 & 52.0 \\
LinkAlign & 66.0 & 48.9 \\
ChessScaler + CHESSSelector + LinkAlignOptimizer & 61.2 & 47.9 \\
ChessScaler + CHESSSelector + MACSQLOptimizer & 68.8 & 42.6 \\
MACSQLScaler + CHESSSelector + LinkAlignOptimizer & 56.5 & 46.8 \\
MACSQLScaler + CHESSSelector + MACSQLOptimizer & 67.3 & 47.9 \\
DINSQLScaler + CHESSSelector + LinkAlignOptimizer & 53.1 & 52.1 \\
DINSQLScaler + CHESSSelector + MACSQLOptimizer & 54.2 & 49.0 \\
RSLSQLScaler + CHESSSelector +LinkAlignOptimizer & 61.7 & 51.0 \\
RSLSQLScaler + CHESSSelector + MACSQLOptimizer & 69.6 & 51.0 \\
RSLSQLScaler + CHESSSelector + DINSQLOptimizer & 66.7 & 50.0 \\
RSLSQLScaler + CHESSSelector + CHESSOptimizer & 66.0 & 51.1 \\
LinkAlignParser + RSLSQLScaler + CHESSSelector + MACSQLOptimizer & 70.8 & / \\
LinkAlignParser + DINSQLScaler + CHESSSelector + MACSQLOptimizer & 68.1 & / \\
LinkAlignParser + ChessScaler + CHESSSelector + MACSQLOptimizer & 68.0 & / \\
LinkAlignParser + MACSQLScaler + CHESSSelector + MACSQLOptimizer & 61.7 & / \\
LinkAlignParser + RSLSQLScaler + CHESSSelector + LinkAlignOptimizer & / & 52.1 \\
LinkAlignParser + DINSQLScaler + CHESSSelector + LinkAlignOptimizer & / & 45.8 \\
LinkAlignParser + ChessScaler + CHESSSelector + LinkAlignOptimizer & / & 55.1 \\
LinkAlignParser + MACSQLScaler + CHESSSelector + LinkAlignOptimizer & / & 51.0 \\
\hline
\end{tabular}
\caption{Comparison of methods on Spider and Bird datasets.}
\label{tab:ex_results}
\end{table*}

\section{Related Work}\label{sec:related_work}

\paratitle{Advances in Text-to-SQL Research.} Early rule-based systems \cite{li2009rule,hong2025survey} relied on hand-crafted templates with limited cross-domain generalization, while neural approaches \cite{choi2021ryansql,hui2021schema} automated NL-to-SQL mapping but struggled with complex queries. Pre-trained language models (PLMs) such as BERT \cite{devlin2019bert} and RoBERTa \cite{liu2019roberta} significantly advanced text-to-SQL through schema-aware encoding \cite{yin2020tabert,li2023graphix,li2023resdsql}. Building on this foundation, Large language models (LLMs) have since driven a paradigm shift: methods like DIN-SQL \cite{pourreza2023dinsql} and DAIL-SQL \cite{gao2024dail} leverage in-context learning to achieve state-of-the-art performance on benchmarks including Spider \cite{yu2018spider} and BIRD \cite{li2023bird}. Recent work introduces collaborative frameworks that coordinate specialized agents for schema linking, SQL generation, and iterative refinement \cite{dong2023c3, zhang2023coder, wang2024macsql}. With the emergence of reasoning strategies such as chain-of-thought (CoT) \cite{wei2022chain} and self-consistency \cite{wang2022self}, advanced methods like CHASE-SQL \cite{pourreza2024chase} further integrate diverse reasoning strategies, boosting performance on complex benchmarks. Building on this
, agentic frameworks including ReFoRCE \cite{deng2025reforce} and LinkAlign \cite{wang2025linkalign} demonstrate strong performance on enterprise-level benchmarks like Spider 2.0 \cite{lei2024spider2} through iterative planning and multi-agent collaboration.

\paratitle{Challenges in Real-World Text-to-SQL Systems.} 
While open-source systems such as DB-GPT \cite{xue2024dbgpt} and DBCopilot \cite{wang2023dbcopilot} have emerged to democratize database access for non-expert users, significant barriers impede their production deployment. First, heterogeneous prompt interfaces and non-standardized model invocation protocols hinder the integration of recent academic advances \cite{zhang2024benchmarking} and result in inference latencies orders of magnitude slower than traditional database systems \cite{kou2024cllms}. Second, robustness analyses reveal systematic vulnerabilities to adversarial perturbations \cite{pi2022adversarial}, lexical variations \cite{gan2021synonym}, and schema incompleteness \cite{deng2021structure}. Third, evaluations on enterprise-grade benchmarks—including Spider 2.0 \cite{lei2024spider2} and BIRD \cite{li2023bird}—demonstrate that state-of-the-art models successfully handle only a small fraction of real-world queries, with diagnostic datasets such as ADVETA \cite{zhong2022adveta} and Dr. Spider \cite{yu2023drspider} further exposing critical gaps in cross-domain generalization. These findings underscore the necessity of unified frameworks that provide standardized interfaces, modular architectures, and enhanced domain adaptability \cite{hong2025survey} to bridge the gap between research and production-ready systems.

\section{Actor Combination Exploration}\label{app:actor_explore}

We summarize the results of all actor combination experiments in Table~\ref{tab:ex_results}. 
Based on these findings, we construct two high-performing multi-actor workflows using these actors as the foundation for subsequent formal experiments.

\section{Runtime Efficiency}\label{app:runtime}

We randomly sample 50 instances from Bird-dev dataset and measure runtime efficiency by using Qwen-turbo as the backbone. As shown in Table~\ref{app:runtime_efficiency}, our reproduced baseline methods maintain efficient performance, with the ensemble variants achieving competitive runtime and moderate token usages despite integrating multiple methods.

\begin{table}[H]
\small
\centering
\begin{tabular}{lcc}
\toprule
\textbf{Methods} & \textbf{Time (s)} & \textbf{Token} \\
\midrule
DIN-SQL      & 7.83  & 9761.5  \\
CHESS-SQL    & 29.46 & 21549.4 \\
MAC-SQL      & 5.47  & 6718.4  \\
RSL-SQL      & 7.34  & 9714.5  \\
LinkAlign    & 31.43 & 18565.8 \\
Ensemble-1   & 33.15 & 20603.5 \\
Ensemble-2   & 28.89 & 10096.5 \\
\bottomrule
\end{tabular}
\caption{Runtime efficiency of different methods.}
\label{app:runtime_efficiency}
\end{table}

\section{Universal Startup Example}\label{app:example}
Squrve provides a universal execution paradigm for Text-to-SQL tasks using a fixed code template and a lightweight configuration file. As illustrated in Figure~\ref{fig:code}, the script initializes a Router to load configurations and an Engine to execute and evaluate tasks. The configuration file specifies all task-related settings, including API credentials, LLM parameters, dataset sources, and task definitions. For instance, to compare DIN-SQL and CHESS on the Spider-dev dataset, users simply define two task entries with their respective generator types (DINSQLGenerator and CHESSGenerator) and specify the execution order in the \emph{exec\_process} field.






\begin{figure}[H]
\centering
\begin{minipage}{\linewidth}
\begin{lstlisting}[language=Python]
from core.base import Router
from core.engine import Engine

if __name__ == "__main__":
    router = Router(config_path="config.json")
    engine = Engine(router)

    # Executing custom tasks...
    engine.execute()

    # Evaluating the results...
    engine.evaluate()

    print("The work is over!")
\end{lstlisting}
\end{minipage}
\caption{Universal execution code of Squrve.}
\label{fig:code}
\end{figure}

\begin{figure}[H]
\small
\centering
\begin{minipage}{0.95\linewidth}
\begin{lstlisting}[numbers=none]
{
  "api_key": {
    "qwen": "your_api_key_here"
  },
  "llm": {
    "use": "qwen",
    "model_name": "qwen-turbo",
    "temperature": 0.75
  },
  "dataset": {
    "data_source": "spider:dev:"
  },
  "database": {
    "schema_source": "spider:dev"
  },
  "task": {
    "task_meta": [
      {
        "task_id": "din_gen",
        "task_type": "GenerateTask",
        "data_source": "spider:dev:",
        "schema_source": "spider:dev",
        "eval_type": ["execute_accuracy"],
        "meta": {
          "task": {
            "generate_type": "DINSQLGenerator"
          }
        }
      },
      {
        "task_id": "chess_gen",
        "task_type": "GenerateTask",
        "data_source": "spider:dev:",
        "schema_source": "spider:dev",
        "eval_type": ["execute_accuracy"],
        "meta": {
          "task": {
            "generate_type": "CHESSGenerator"
          }
        }
      }
    ]
  },
  "engine": {
    "exec_process": ["din_gen", "chess_gen"]
  }
}
\end{lstlisting}
\end{minipage}
\caption{A simple configuration file for comparing DIN-SQL and CHESS on the Spider-dev dataset.}
\label{fig:config}
\end{figure}

\end{document}